\documentclass[11pt]{article}

\usepackage[final]{acl}

\usepackage{times}
\usepackage{latexsym}
\usepackage[T1]{fontenc}
\usepackage[utf8]{inputenc}
\usepackage{microtype}
\usepackage{inconsolata}
\usepackage{graphicx}
\usepackage{booktabs}
\usepackage{tabularx}
\usepackage{array}
\usepackage{placeins}
\usepackage{xcolor}
\usepackage{amssymb}
\usepackage{enumitem}

\newcommand{\benchmark}{\textsc{TSW-NL}}
\newcommand{\pass}{\textcolor{green!45!black}{\(\checkmark\)}}
\newcommand{\fail}{\textcolor{red!70!black}{\(\times\)}}
\newcommand{\mixed}{\textcolor{orange!85!black}{\(\sim\)}}
\newcommand{\para}[1]{\vspace{-0.05cm}\paragraph{#1}\vspace{-0.05cm}}
\newcommand{\paras}[1]{\vspace{-0.05cm}\paragraph{#1}\vspace{-0.05cm}}
\title{The Shared Word: Evaluating Language Models on Dutch Word Puzzles}

\author{Rik van Noord \and Maurice Voors   \\
  CLCG, University of Groningen \\
  \texttt{r.i.k.van.noord@rug.nl, mauricevoors@hotmail.com}}

\begin{document}
\maketitle

\begin{abstract}
Many LLM benchmarks for languages other than English are translated, multiple-choice, or difficult to score automatically. We introduce \emph{The Shared Word} (\benchmark{}), a Dutch benchmark built around four core principles: it is language-specific, open-vocabulary, human-curated, and deterministically scored. Each puzzle gives three clues that describe different senses or uses of the same Dutch word. The clues are individually ambiguous, requiring the solver to combine evidence across multiple clues. We construct the benchmark from Dutch Wikipedia disambiguation pages, followed by automatic filtering and extensive manual curation. The final dataset contains 222 instances, with three independent human answers per test item. The task is clearly challenging: scores in our human study range from 49.5\% to 61.0\% and the highest-scoring open-weight model we tested reaches only 39.3\%. Many others even score below 20\%. At the same time, closed-source models show great progress: GPT-4.1 (April 2025) scores 59.2\%, whereas GPT-5.6 Sol with max reasoning scores 88.8\%. We believe this task provides a useful addition to Dutch model evaluation, plus a clear recipe that can be applied to construct a similar benchmark for other languages.
\end{abstract}

\section{Introduction}
Reliable benchmarking is important for understanding what large language models (LLMs) can and cannot do, in particular for languages other than English. Many of the benchmarks commonly used for such languages are directly translated from English, making it unclear whether LLMs are truly capable in those languages. Similarly, many tasks are multiple-choice, which only test whether the models recognize the correct answer, instead of generating it themselves. Moreover, tasks that require generation often have unclear and fuzzy evaluation metrics, making comparison between models hard.

\begin{figure}[t]
    \centering
    \includegraphics[width=0.8\columnwidth]{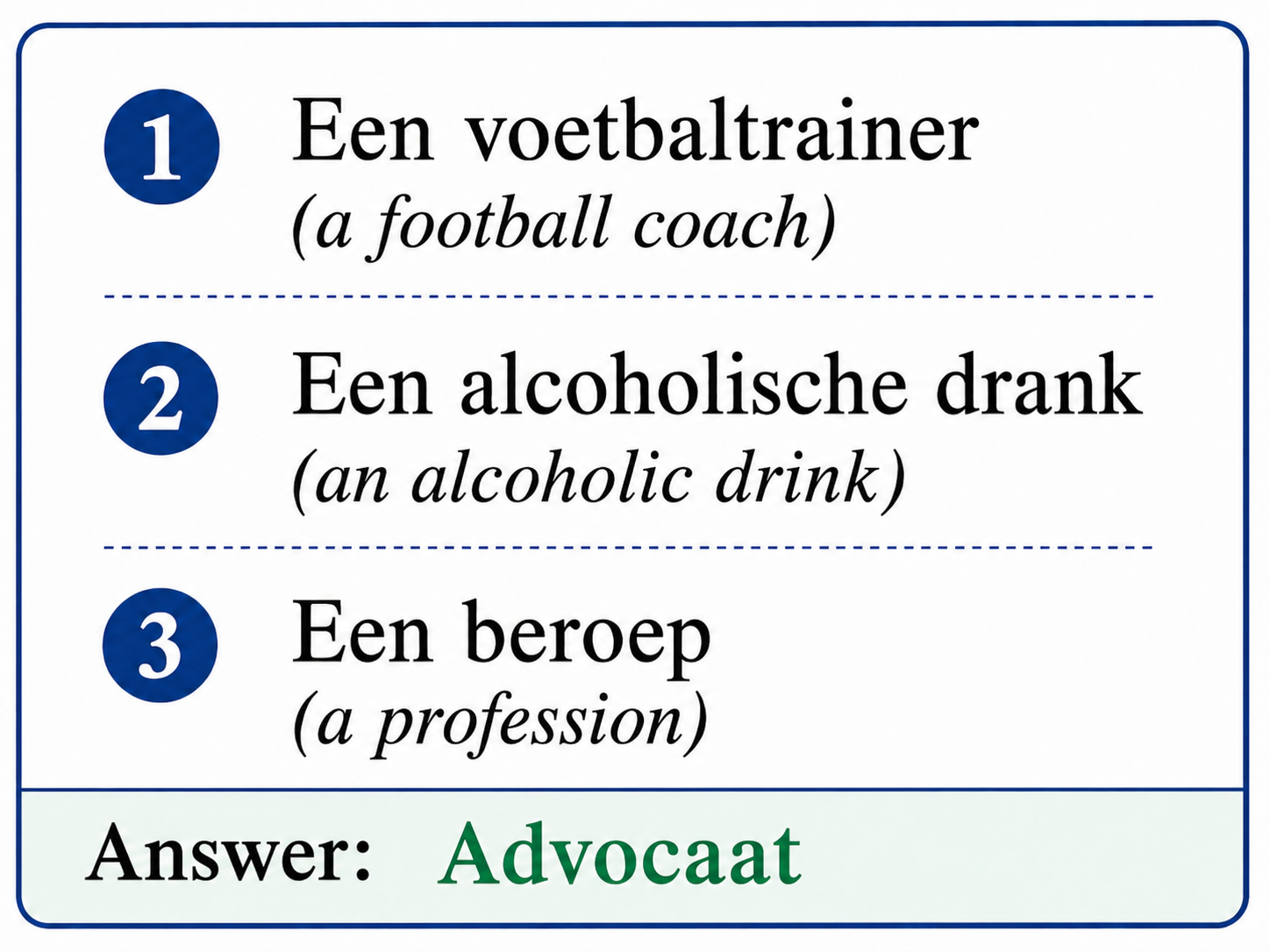}
    \caption{Example puzzle. The three Dutch clues describe a football coach (Dick Advocaat), an alcoholic drink (Dutch-specific, similar to eggnog), and a profession (lawyer), all answered by \emph{advocaat}.
    }
    \vspace{0.1cm}
    \label{fig:advocaat-example}
\end{figure}

To address these issues, we introduce a Dutch benchmark based on four simple principles:

\begin{enumerate}
\item The benchmark is language-specific: the lexical relation between its clues and answer does not survive direct translation.
\item The benchmark requires open-vocabulary generation.
\item Each item has one short intended answer and can be scored deterministically.
\item All items in the benchmark are human-verified and curated.
\end{enumerate}

This may sound straightforward, but finding benchmarks that meet all four requirements is surprisingly difficult.\footnote{See Appendix Tables~\ref{tab:related-benchmarks}--\ref{tab:dutch-benchmark-tasks} for a detailed comparison.}

\paragraph{Task format}
The task we introduce consists of three clues that all point to a single Dutch word. Figure~\ref{fig:advocaat-example} shows an example. It is inspired by a popular question type in the Dutch TV quiz \emph{Met Het Mes Op Tafel}. There are no answer options or predefined labels. The model must interpret the clues, combine them, and retrieve the word from an open vocabulary. The clues are curated around one intended answer and they are deliberately underspecified as not to point to only a single term. We use disambiguation pages of Dutch Wikipedia to create our dataset. After extensive filtering, we manually curated it to ensure, among other things, that the three clues collectively admit only one intended answer, the examples cannot be translated while preserving the same answer, and the clues are distinctly different and do not contain obscure words or references. Importantly, we want a task that is still doable for humans, so the LLM capabilities can be evaluated in the light of human performance as well.

\para{Dutch benchmarks} Dutch already has strong resources: DUMB combines nine language-understanding tasks, the EuroEval Natural Language Understanding benchmark includes Dutch, and BEIR-NL tests information retrieval \citep{de-vries-etal-2023-dumb,saattrup-nielsen-etal-2025-encoder,lotfi-etal-2025-beir}. What our work adds is a difficult free-generation task that is easy to score. We also test whether models adapted to Dutch, such as GEITje, Fietje 2 and ChocoLlama \citep{vanroy-2024-geitje,vanroy-2024-fietje,meeus-etal-2024-chocollama}, perform relatively better on a Dutch-specific benchmark such as this one.

\para{Results}

The final, heavily curated test set contains 200 instances, each answered by three human annotators. Experiments show that the benchmark has a useful level of difficulty. All evaluated open-weight models struggle, given that the strongest approach only reaches 39.3\% accuracy. There are still clear differences between the models, indicating that our benchmark seems to distinguish well between different model capabilities. The strongest closed models perform much better: they clearly exceed the scores of our human annotators, while still being quite far from perfect.

\para{Contributions}
While the benchmark itself is language-specific, the process that creates it is not. Our main contribution is therefore a novel NLP task for testing language-specific lexical knowledge, candidate retrieval, and evidence integration through an open-vocabulary task that does not transfer directly across languages. In particular, we release \benchmark{}, a carefully curated set of 222 Dutch puzzle instances that serves as a useful addition to current Dutch LLM evaluation benchmarks.

\begin{table*}[t]
\centering
\footnotesize
\begin{tabularx}{\textwidth}{p{4.5cm}rrX}
\toprule
\textbf{Stage} & \textbf{Left} & \textbf{Removed} & \textbf{Operation} \\
\midrule
\multicolumn{4}{l}{\textbf{Automatic collection and filtering}} \\
\addlinespace[2pt]
Dutch Wikipedia disamb. pages & 97,707 & 0 & Retrieve candidate answer titles. \\
At least three linked pages & 48,518 & 49,189 & Require enough links for three clues. \\
Title in Open Dutch WordNet & 5,723 & 42,795 & Prefer lexical answers over proper-name-only pages. \\
Title has at least four characters & 5,075 & 648 & Remove short abbreviations and acronyms. \\
At least three described linked pages & 4,159 & 916 & Require three links with descriptions and categories. \\
Informative descriptions & 4,047 & 112 & Remove generic, Wikipedia-internal, and numeric clues. \\
Non-media category mix & 3,057 & 990 & Remove candidates dominated by media and place categories. \\
No answer leakage & 2,929 & 128 & Remove descriptions containing the answer. \\
Distinct descriptions & 2,896 & 33 & Remove repeated descriptions. \\
Sufficiently viewed pages & 2,479 & 417 & Remove pages with fewer than 3,000 views in the previous month. \\
No unsuitable domain descriptions & 2,087 & 392 & Remove formulaic location, taxon, building, and related clues. \\
\addlinespace[2pt]
\midrule
\multicolumn{4}{l}{\textbf{Manual annotation and selection}} \\
\addlinespace[2pt]
Obscurity \& sense & 568 & 1,519 & Remove obscure candidates and check for distinct senses. \\
Final quality selection & 222 & 346 & Check for a single answer, the need for multiple clues, and Dutch specificity. \\
\bottomrule
\end{tabularx}
\vspace{-0.1cm}
\caption{Data reduction during automatic filtering and manual annotation.}
\vspace{-0.15cm}
\label{tab:pipeline}
\end{table*}

\section{Related Work}

\para{Relation to existing tasks}
In a sense, our benchmark runs word-sense disambiguation in reverse \citep{navigli-2009-wsd}. Instead of mapping a given word to a fixed sense inventory, the solver recovers the word from several senses or conventional uses. It also resembles riddle and lateral-reasoning benchmarks \citep{lin-etal-2021-riddlesense,jiang-etal-2023-brainteaser,manna-etal-2024-riddle,samdarshi-etal-2024-connecting,saha-etal-2025-language,gevers-daelemans-2026-get}, but is language-specific and has no answer choices. Solving it requires lexical knowledge, candidate search, and evidence integration. The retrieval direction is close to the reverse dictionary task, where a description is used to find a word \citep{hill-etal-2016-learning-understand,pilehvar-2019-importance,yan-etal-2020-bert,qi-etal-2020-wantwords}. 
However, our task is quite different. For one, reverse dictionary systems typically recover a word from a description of just one sense. Moreover, here each clue is deliberately underspecified and describes a different sense. Potential solvers must therefore combine candidates across clues, with the shared word form often being their only connection. Wikipedia disambiguation pages support this construction well, since they group ordinary senses with people, places, organizations, titles, and other cultural referents. They also allow us to remove obscure references by checking page views.

\para{Language-specific benchmarks}
We construct the task in Dutch rather than translate it from English. Translation is useful for cross-lingual comparison, but translation quality and artefacts can affect multilingual benchmark results \citep{thellmann-etal-2024-european}. The benchmark does not transfer directly across languages: senses that share a form in Dutch generally use unrelated words in another language. There are several recent benchmarks that are also language or culture specific, such as Korean social values, Persian etiquette and customs, Arabic financial reasoning, and Turkish temporal interpretation \citep{lee-etal-2024-kornat,sadr-etal-2025-politely,sakhaeirad-etal-2026-unmasking,elbadry-etal-2026-sahm,demir-etal-2026-timeres}. There are, however, not many that follow our four principles as described in the introduction. Appendix Tables~\ref{tab:related-benchmarks} and~\ref{tab:related-benchmark-suites} compare a very broad set on these principles and find that only a few tasks in larger language-specific benchmarks pass all four checks.\footnote{We extensively searched for benchmarks and evaluated what we found, but there are many benchmarks these days and we surely missed some.}
Note that this is not to say that four principles are the only important qualities for a benchmark. There are many more important qualities. But it does highlight that our benchmark possesses interesting qualities that are not common.

\para{Dutch benchmarks}
There are already a few well-known Dutch LLM benchmarks. DUMB \citep{de-vries-etal-2023-dumb} evaluates Dutch language understanding through nine tasks, including part-of-speech tagging, named-entity recognition, word-sense discrimination, pronoun resolution, inference, sentiment, abusive-language classification, and question answering. EuroEval \citep{saattrup-nielsen-etal-2025-encoder} places Dutch in a broader European evaluation suite and covers classification, extraction, summarization, knowledge, common-sense reasoning, and simplification. BEIR-NL adds a task on Dutch information retrieval \citep{lotfi-etal-2025-beir}.
Appendix Table~\ref{tab:dutch-benchmark-tasks} shows that none of the existing tasks pass all four principles, meaning that we believe our task would be a valuable addition to the community.

\section{Benchmark Construction}
\label{sec:benchmark}

\para{Task}
As shown in Figure~\ref{fig:advocaat-example}, each puzzle has three Dutch clues and one Dutch answer. Each clue describes a different sense or referent of the answer word. Together they should point to one intended answer. We design the clues so that no single clue should identify the answer on its own. Their combination should narrow the possibilities to the intended answer.
For example, in Figure~\ref{fig:advocaat-example}, clues 1 and 2 could also point to \emph{Rose}, clues 1 and 3 to \emph{Boer}, and clues 2 and 3 to \emph{Kardinaal}.\footnote{The football coaches are Marco Rose and Frank de Boer.}
Appendix Table~\ref{tab:test-examples} shows ten more puzzle examples.

\subsection{Filtering and Curation}

\para{Source and initial filters}
We start by collecting all Dutch Wikipedia disambiguation pages.\footnote{For example, the corresponding page for the puzzle in Figure~\ref{fig:advocaat-example} is \url{https://nl.wikipedia.org/wiki/Advocaat}.} The page title then is a candidate answer and descriptions of linked pages are candidate clues. The initial collection has 97,707 pages. We require at least three links (since we want three clues) and remove all short abbreviations. We also only keep titles found in Open Dutch WordNet \citep{postma-etal-2016-open}: we want all the answers to be common Dutch words. This ensures that we remove instances for which all the clues point to a given name.\footnote{E.g., this way there are no puzzles pointing to 3 pages about people with the common Dutch last name \emph{De Jong}.} Moreover, the puzzles are most interesting when there is a mix of clues. We also filter pages where the clues are not descriptive enough (e.g., too short or numerical).

\para{Detailed filters} At this point (4,047 examples left), a manual inspection showed us that the benchmark still has a few problems. For one, some puzzles are too easy, and a single clue often just gives away the answer. On the other hand, some puzzles are simply way too hard, as they for example reference an obscure Dutch city or person that almost no person realistically will have heard of. Moreover, clues are sometimes too similar, which does not lead to an engaging puzzle.\footnote{For example, a band may release an album with the same title as the band.} Therefore, we apply a few more filters to only keep interesting and engaging puzzles. We filter puzzles by media and place categories, as they overwhelmingly reference obscure places and titles. We also exclude puzzles in which one of the clue descriptions contains the answer directly, and remove repeated descriptions. Then, we filter puzzles that have clues that are too formulaic and Wikipedia-specific, such as "region of", "taxon", and "building in". Finally, as an extra filter for too obscure answers, we filter linked pages below 3,000 views in the previous month.

\para{Manual curation}
At this point (see Table~\ref{tab:pipeline}), we have 2,087 puzzles left, but we can tell there are still clear issues with the resulting puzzles that automatic filtering cannot capture. We turn to manual filtering, with one primary curator, who discussed interesting cases with a second curator. First, we filter all instances where any one of the clues is simply deemed too obscure, or where two clues effectively point to the same sense. This is subjective to some extent: when is something obscure and when is it not? Nevertheless, this is a necessary step, and we trust humans a lot more on this task than LLMs or other automated methods.

\para{Final manual checks} We then have 568 potential puzzles left, but there are still a few more checks to do. For one, we only want puzzles for which all three clues are ideally needed to arrive at the final answer. We therefore filter examples where clearly only one or two clues would be needed to solve the puzzle. This works well, but is also a fine line: we see later in Table~\ref{tab:ablation-results} that models can still solve some of the puzzles when only seeing one or two clues. Second, we want the puzzle to be completely Dutch-specific. Third, we carefully check that the clues point to one intended answer and that no obvious alternative remains.\footnote{Appendix Table~\ref{tab:plausible-alternatives} lists seven cases that still got past us, detected during error analysis. These alternatives were added as acceptable answers.}

\para{Rewriting}
For the puzzles that pass this check, we manually rewrite the clues.
Wikipedia descriptions were written to identify pages, not to make puzzles. Some contain dates, locations, or parenthetical details that give away too much information, while others are too vague outside the context of the page. We shorten these clues or add just enough context to make them readable, while keeping the underlying sense unchanged.
The final dataset of 222 instances (randomly split into 22 development and 200 test instances) consists entirely of manually reviewed and curated items.
\footnote{Data and code are available at \nolinkurl{https://github.com/RikVN/TSW-NL}. Licensing details in Appendix~\ref{app:licenses}.}

\para{Size}
The test set contains 200 carefully curated items. This is relatively small for a benchmark. We prioritize item quality over a larger set containing weaker puzzles. To reduce dependence on clue order, each model is evaluated on all six permutations, producing 1,200 predictions while treating the six predictions for each item as repeated measurements. Also, as Table~\ref{tab:main-results} will later show, the resulting scores still reveal substantial differences between models.

\para{Dataset characteristics}
The items mix lexical senses with names and cultural references. We manually reviewed all 600 test clues and counted the clues tied to a specifically Dutch entity or cultural context. We found 47 such clues (7.8\%) across 38 test puzzles (19.0\%). Note that the rest of the benchmark is still Dutch-specific through its lexical ambiguity. Across the dataset, models solving this task must handle polysemy, homonymy, metonymic or conventional uses, names of entities, titles, brands or technical senses. These categories can also overlap within an item.

\para{Wikipedia exposure}

Because the candidate senses originate from Dutch Wikipedia, pretrained models may benefit from associations acquired from that source, and models may differ in the extent of their exposure. However, all systems are evaluated closed-book, and every open-weight model scores below 40\%, indicating that source exposure alone is clearly insufficient to solve the task reliably.

\subsection{Human Evaluation}

To get a good sense of human performance, we asked nine annotators to attempt to solve the puzzles, such that each puzzle item had three attempted answers. Seven Dutch speakers answered blocks of 20 or 40 items, together covering the test set once. Two more annotators each answered all 200 items. All nine participants are native speakers and took part voluntarily. The full annotation protocol is available in Appendix~\ref{app:human}.
It turns out that human annotators can do the task quite well, but they are also far from perfect. The pooled block of 7 annotators contains 103 correct answers (51.5\%). The two full-test annotators solved 99 (49.5\%) and 122 items (61.0\%). 
There are still 43 items that no human managed to solve. This confirms that the benchmark remains challenging despite the strict manual filtering for obscure clues.

\section{Experimental Setup}
\label{sec:setup}

To see how well LLMs do on this benchmark, we test a large number of models. We test instruction-tuned models in a closed-book setting, without retrieval or any Wikipedia or internet access. We define the task as zero-shot: a Dutch prompt explains that the clues point to one word and asks for only that word.\footnote{We ran some preliminary experiments in which we explained the task in English instead, and usually obtained slightly worse performance.} The exact prompts are available in Appendix~\ref{app:prompts}.

\vspace{0.05cm}
\paragraph{Models}
\vspace{0.05cm}
We cover Dutch-specific models and multilingual models (with and without reasoning) of varying sizes and families. The benchmark is intended to help open-weight developers evaluate their models, but we understand the curiosity to see how well state-of-the-art closed-source models do. Therefore, we test GPT-5.6 Luna and Sol and include two earlier models from around the time work on this benchmark began, to show progress. Table~\ref{tab:models} lists every model and groups them by model type.
To investigate whether further training on Dutch helped the Dutch-specific models, we add the closest instruction-tuned controls for two lineages: Mistral~7B Instruct v0.1 for GEITje 7B Ultra and Llama~3 8B Instruct for ChocoLlama 3 8B Instruct.

\vspace{0.05cm}
\paragraph{Prompting conditions}
\vspace{0.05cm}
We design the task as zero-shot: the clues should contain enough information without worked examples. Out of interest, we compared performance using 0, 1, 3, 5, 10, and 22 examples in canonical clue order for a selection of our models, shown in Appendix Table~\ref{tab:shot-selection}. Differences are generally small, and some models in fact do best with no examples. For the reasoning models, we use the highest reasoning setting available.

\begin{table}[t]
\vspace{0.4cm}
\centering
\scriptsize
\setlength{\tabcolsep}{2pt}
\resizebox{\columnwidth}{!}{%
\begin{tabular}{@{}l r l l@{}}
\toprule
\textbf{Model} & \textbf{Size} & \textbf{Type} & \textbf{Reference} \\
\midrule
Fietje 2 Instruct & 2.7B & Dutch & \citealp{vanroy-2024-fietje} \\
Fietje 2 Chat & 2.7B & Dutch & \citealp{vanroy-2024-fietje} \\
GEITje 7B Ultra & 7B & Dutch & \citealp{vanroy-2024-geitje} \\
ChocoLlama 3 8B Instruct & 8B & Dutch & \citealp{meeus-etal-2024-chocollama} \\
\midrule
Mistral 7B Instruct v0.1 & 7B & Multilingual & \citealp{jiang-etal-2023-mistral} \\
Llama 3 8B Instruct & 8B & Multilingual & \citealp{dubey-etal-2024-llama3} \\
Salamandra 7B Instruct & 7B & Multilingual & \citealp{gonzalez-agirre-etal-2025-salamandra} \\
Aya Expanse 8B & 8B & Multilingual & \citealp{dang-etal-2024-aya-expanse} \\
Aya Expanse 32B & 32B & Multilingual & \citealp{dang-etal-2024-aya-expanse} \\
Llama 3.1 8B Instruct & 8B & Multilingual & \citealp{dubey-etal-2024-llama3} \\
Gemma 3 4B IT & 4B & Multilingual & \citealp{gemma-team-2025-gemma3} \\
Gemma 3 12B IT & 12B & Multilingual & \citealp{gemma-team-2025-gemma3} \\
Gemma 3 27B IT & 27B & Multilingual & \citealp{gemma-team-2025-gemma3} \\
EuroLLM-9B Instruct & 9B & Multilingual & \citealp{martins-etal-2025-eurollm9b} \\
EuroLLM-22B Instruct & 22B & Multilingual & \citealp{ramos-etal-2026-eurollm22b} \\
Ministral 3 14B Instruct & 14B & Multilingual & \citealp{mistral-ai-2025-ministral3-14b} \\
\midrule
Qwen3-8B & 8B & Multi/reason & \citealp{yang-etal-2025-qwen3} \\
Qwen3.5-9B & 9B & Multi/reason & \citealp{qwen-team-2026-qwen35-9b} \\
Phi-4 Mini Reasoning & 3.8B & Multi/reason & \citealp{xu-etal-2025-phi4-mini-reasoning} \\
Phi-4 Reasoning & 14B & Multi/reason & \citealp{abdin-etal-2025-phi4-reasoning} \\
GPT-OSS-20B & 20B & Multi/reason & \citealp{openai-2025-gpt-oss-card} \\
\midrule
GPT-4o (2024-11-20) & -- & Multilingual & \citealp{openai-2026-model-docs} \\
GPT-4.1 (2025-04-14) & -- & Multilingual & \citealp{openai-2026-model-docs} \\
GPT-5.6 Luna & -- & Multi/reason & \citealp{openai-2026-model-docs} \\
GPT-5.6 Sol & -- & Multi/reason & \citealp{openai-2026-model-docs} \\
\bottomrule
\end{tabular}%
}
\vspace{0.2cm}
\caption{Models included in the evaluation.}
\label{tab:models}
\vspace{0.1cm}
\end{table}

\vspace{0.05cm}
\paragraph{Scoring}
\vspace{0.05cm}
Our main metric is exact-match accuracy after trimming whitespace and lowercasing. 
This is deterministic and requires no subjective or model-based judge.
In our main experiment, every model sees all six clue orders to reduce dependence on any single clue order. We report their mean accuracy and sample standard deviation, which provides an immediate indication of sensitivity to clue order.
We average correctness over the six orders for each item. We obtain 95\% confidence intervals by bootstrapping the 200 items. For model comparisons, we use paired sign-flip tests on these item-level scores and paired bootstrap intervals for the accuracy difference, with Holm correction across the family of comparisons. We test clue-order effects by randomly permuting the six order labels within each item. The iteration counts and further implementation details are available in Appendix~\ref{app:reproducibility}.

\paragraph{Source-retrieval diagnostic}
\vspace{0.05cm}
To further judge the results of the LLMs, we also run a BM25 source-retrieval diagnostic~\citep{robertson-zaragoza-2009-bm25}. Note that this is not a fair baseline, as it receives candidate inventories and Wikipedia source material, but it makes for an interesting comparison. Each candidate answer becomes a document containing its linked senses, the three clues form the query, and the title of the highest-ranked document is returned. We use two source-matched pools with linked-page titles and Dutch descriptions: all 2,087 automatically filtered candidates and, as an oracle-like condition, only the 200 test answers. This diagnostic basically asks how directly the clues match their source descriptions and how plausible recovery from memorized source associations would be. BM25 reaches 69.0\% with the 2,087-candidate pool and 85.5\% with the oracle-like 200-answer pool. This shows that the clues do contain substantial source signal, though it is also far from enough to completely solve the task.

\subsection{Error annotation}
To understand the remaining errors, we manually annotate every answer initially marked incorrect for a subset of models and one human annotator. 
We assign each response to one of the following mutually exclusive categories:

\begin{itemize}[itemsep=1pt]
    \item \textbf{Empty}: the human left the answer blank, or the model produced no final answer before reaching its token limit.
    \item \textbf{Repeats (part of) a clue}: the answer copies wording from a clue instead of identifying the shared word.
    \item \textbf{Completely incorrect}: none of the other categories apply and the answer does not meaningfully fit the clues.
    \item \textbf{Fits only one clue}: the answer is related to one clue but not the other two.
    \item \textbf{Fits only two clues}: the answer combines part of the evidence but fails on the third clue.
    \item \textbf{Inflection variant}: the answer is morphologically related to an accepted answer, but the produced form does not itself satisfy all three clues.
    \item \textbf{Correct alternative}: another answer that also fits all three clues.
\end{itemize}
 
This analysis also made us realize that checking wrong answers for only a few models is unfair. However, annotating all 18,866 unique non-empty wrong responses across all experiments is not feasible. Therefore, we additionally annotate every previously unreviewed answer that occurs at least six times across the experiments, but only to determine whether the answer should be accepted.\footnote{There are six clue orders, so an annotated answer could come from a single model.}
Of these 532 unique answers, only three should have been accepted. Appendix Table~\ref{tab:plausible-alternatives} lists all these accepted alternatives, together with the four unique ones found in the more extensive manual annotation (Table~\ref{tab:error-annotation-categories-main}). We then rescore all experiments with the expanded answer key, e.g. allowing some puzzles to have multiple answers. This produces only small changes, but is in our opinion the most fair setup.

\begin{table}[t]
\small
\centering
\begin{tabular}{lrr}
\toprule
\textbf{System} & \textbf{Acc.} & \textbf{SD} \\
\midrule
Human annotators & 54.0 & \multicolumn{1}{c}{---} \\
\midrule
\textbf{Dutch models} & & \\
\hspace{0.15cm} Fietje 2 Instruct & 7.0 & 1.0 \\
\hspace{0.15cm} Fietje 2 Chat & 9.3 & 0.8 \\
\hspace{0.15cm} GEITje 7B Ultra & 14.8 & 2.1 \\
\hspace{0.15cm} ChocoLlama 3 8B Instruct & 15.0 & 1.7 \\
\midrule
\textbf{Open-weight non-reasoning} & & \\
\hspace{0.15cm} Mistral 7B Instruct v0.1 & 5.1 & 1.3 \\
\hspace{0.15cm} Llama 3 8B Instruct & 14.2 & 2.8 \\
\hspace{0.15cm} Salamandra 7B Instruct & 13.1 & 2.8 \\
\hspace{0.15cm} Aya Expanse 8B & 14.0 & 1.0 \\
\hspace{0.15cm} Aya Expanse 32B & 32.9 & 1.2 \\
\hspace{0.15cm} Llama 3.1 8B Instruct & 15.8 & 0.9 \\
\hspace{0.15cm} Gemma 3 4B IT & 15.6 & 2.0 \\
\hspace{0.15cm} Gemma 3 12B IT & 29.8 & 1.4 \\
\hspace{0.15cm} Gemma 3 27B IT & 39.3 & 1.5 \\
\hspace{0.15cm} EuroLLM-9B Instruct & 25.8 & 1.5 \\
\hspace{0.15cm} EuroLLM-22B Instruct & 27.3 & 1.8 \\
\hspace{0.15cm} Ministral 3 14B Instruct & 30.8 & 0.7 \\
\midrule
\textbf{Open-weight reasoning} & & \\
\hspace{0.15cm} Qwen3-8B & 28.5 & 1.5 \\
\hspace{0.15cm} Qwen3.5-9B & 19.9 & 2.6 \\
\hspace{0.15cm} Phi-4 Mini Reasoning & 10.3 & 2.0 \\
\hspace{0.15cm} Phi-4 Reasoning & 9.0 & 1.3 \\
\hspace{0.15cm} GPT-OSS-20B & 12.1 & 1.2 \\
\midrule
\textbf{Proprietary models} & & \\
\hspace{0.15cm} GPT-4o (2024-11-20) & 56.3 & 2.5 \\
\hspace{0.15cm} GPT-4.1 (2025-04-14) & 59.2 & 0.7 \\
\hspace{0.15cm} GPT-5.6 Luna (none) & 54.1 & 1.4 \\
\hspace{0.15cm} GPT-5.6 Luna (max) & 79.0 & 1.3 \\
\hspace{0.15cm} GPT-5.6 Sol (none) & 69.9 & 0.9 \\
\hspace{0.15cm} GPT-5.6 Sol (max) & 88.8 & 1.2 \\
\bottomrule
\end{tabular}
\caption{Test accuracy (\%), averaged over the six permutations of the three clues. SD is the sample standard deviation of the six order-level accuracies.}
\label{tab:main-results}
\vspace{-0.2cm}
\end{table}

\section{Results}
\label{sec:results}

Table~\ref{tab:main-results} shows the accuracy scores for the LLMs used in this study.\footnote{Appendix Figure~\ref{fig:main-results-forest} shows the same results, but now visualized and with 95\% confidence intervals.}
Scores range from 5.1\% (Mistral 7B) to an impressive 88.8\% for GPT-5.6 Sol. Generally, all models except the very best ones clearly struggle with the task. We are happy to see that, as it shows that the benchmark is far from saturated, at least for open-weight and non-reasoning models. 
Among the open-weight models, Gemma~3 27B scores highest at 39.3\%, followed by Aya Expanse 32B at 32.9\% and Ministral~3 14B at 30.8\%.
Larger models usually do better than their smaller counterparts, except for the EuroLLM models, which do not differ significantly in score. The Dutch-specific models unfortunately do not perform as well, though it is interesting to see that GEITje and ChocoLlama perform better than \mbox{Fietje~2}.

\paras{Significance} The statistical tests support the clearest differences in the table. GPT-5.6 Sol outperforms every other model after Holm correction, while Gemma~3~27B is significantly better than all other open-weight models. In general, we find that clue order has little systematic effect. Of the 27 model conditions shown in Figure~\ref{fig:main-results-forest}, 25 have an uncorrected randomization $p>.10$.\footnote{Llama~3 8B ($p=.011$) and Salamandra 7B ($p=.012$) are the exceptions, but they do not survive Holm correction.}

\paras{Does further Dutch training help?}

To answer this, we compare the GEITje and ChocoLlama models to the two parent models they were based on. We cannot make the same comparison for Fietje 2: its base model, Phi-2, is not instruction-tuned and could not perform the prompted task at all. The picture is mixed. For GEITje it clearly helped, as it reaches 14.8\%, compared with 5.1\% for its parent model Mistral~7B Instruct.
However, ChocoLlama reaches 15.0\%, compared with 14.2\% for its parent model Llama~3 8B Instruct, which is not a significant improvement.

\paras{Does reasoning help?}

Allowing reasoning makes a large difference for both GPT-5.6 models (see the bottom of Table~\ref{tab:main-results}). Without it, Luna drops by 24.9 points and Sol by 18.9 points. Even though this is not really a ``reasoning'' task in the traditional sense, the models clearly benefit from being able to spend more compute on thinking through and evaluating potential answers.

\paras{How much did SOTA models improve?}
The dated GPT snapshots at the bottom of Table~\ref{tab:main-results} give a rough view of progress: GPT-4o from November 2024 scores 56.3\%, GPT-4.1 from April 2025 scores 59.2\%, and GPT-5.6 Sol (July 2026) with max reasoning reaches 88.8\%. Of course this is not a controlled architecture comparison, but the 29.6-point difference between GPT-4.1 and Sol shows how much performance on this task changed over roughly fifteen months.

\begin{figure}[t]
\vspace{0.3cm}
\centering
\includegraphics[width=0.96\columnwidth]{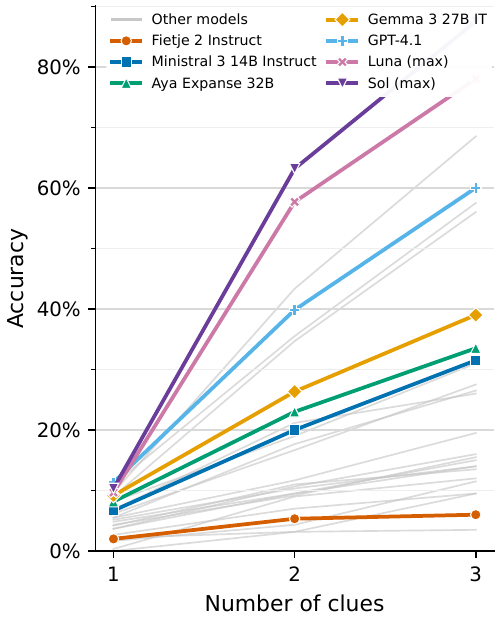}
\vspace{0.3cm}
\caption{Model accuracy by number of clues. For clarity, only a few models are highlighted in color. The other models are shown in grey. One- and two-clue scores average over the three corresponding clue subsets.}
\label{fig:clue-ablation}
\vspace{0.2cm}
\end{figure}

\begin{table}[t]
\centering
\small
\begin{tabular}{lrrrr}
\toprule
\textbf{Model} & \textbf{0/3} & \textbf{1/3} & \textbf{2/3} & \textbf{3/3} \\
\midrule
Fietje 2 Instruct & 4.3 & 7.4 & 5.0 & 10.5 \\
Fietje 2 Chat & 4.7 & 11.3 & 6.6 & 13.5 \\
GEITje 7B Ultra & 5.0 & 8.9 & 17.0 & 25.1 \\
ChocoLlama 3 8B Instruct & 5.4 & 12.1 & 15.1 & 24.6 \\
\midrule
Mistral 7B Instruct v0.1 & 3.5 & 6.4 & 1.9 & 8.2 \\
Llama 3 8B Instruct & 10.1 & 12.8 & 11.0 & 21.3 \\
Salamandra 7B Instruct & 3.9 & 10.3 & 19.2 & 16.7 \\
Aya Expanse 8B & 9.3 & 12.1 & 6.9 & 25.7 \\
Aya Expanse 32B & 17.8 & 25.5 & 36.8 & 46.8 \\
Llama 3.1 8B Instruct & 7.8 & 13.1 & 14.8 & 24.9 \\
Gemma 3 4B IT & 5.0 & 16.3 & 14.8 & 23.7 \\
Gemma 3 12B IT & 12.8 & 32.3 & 29.6 & 40.9 \\
Gemma 3 27B IT & 21.3 & 33.3 & 41.5 & 55.8 \\
EuroLLM-9B Instruct & 10.1 & 22.3 & 27.0 & 39.5 \\
EuroLLM-22B Instruct & 11.2 & 27.7 & 22.3 & 43.9 \\
Ministral 3 14B Instruct & 15.1 & 27.0 & 29.9 & 46.5 \\
\midrule
Qwen3-8B & 12.0 & 23.8 & 32.1 & 41.5 \\
Qwen3.5-9B & 9.7 & 11.7 & 24.2 & 30.4 \\
Phi-4 Mini Reasoning & 4.7 & 10.6 & 10.4 & 14.3 \\
Phi-4 Reasoning & 2.7 & 6.7 & 9.4 & 15.2 \\
GPT-OSS-20B & 5.8 & 12.4 & 8.5 & 19.9 \\
\midrule
GPT-4o (2024-11-20) & 31.8 & 45.4 & 62.6 & 78.1 \\
GPT-4.1 (2025-04-14) & 31.4 & 50.0 & 68.6 & 78.9 \\
GPT-5.6 Luna (none) & 34.9 & 44.0 & 59.1 & 72.2 \\
GPT-5.6 Luna (max) & 63.2 & 72.0 & 82.7 & 93.3 \\
GPT-5.6 Sol (none) & 46.5 & 63.5 & 75.8 & 87.4 \\
GPT-5.6 Sol (max) & 72.5 & 90.4 & 93.4 & 95.3 \\
\bottomrule
\end{tabular}
\caption{Accuracy (\%) by the number of correct human judgments, averaged over six clue orders. The four groups contain 43, 47, 53, and 57 items.}
\label{tab:human-difficulty-main}
\end{table}

\paras{Do models need all clues?}

We created this dataset with the goal that each puzzle could only be reliably solved when combining information from multiple clues, preferably even all three clues. However, in practice models could obviously still get to the right answer by looking at only two clues, or making an educated guess after only one. This ablation experiment is shown in Figure~\ref{fig:clue-ablation}.\footnote{Complete scores are available in Appendix Table~\ref{tab:ablation-results}.} We see that having three clues is certainly beneficial, but that models can do reasonably well with two. GPT-5.6 Sol (max) averages only 10.3\% with one clue, already 63.2\% with two, and all three  get the model close to 90\%. Most open-weight checkpoints in our selection also improve with more evidence.
One clue is almost never enough (apart from some lucky guesses), but two clues already help the best models a lot. In future work, it would be interesting to run a similar experiment with human annotators, to see if it helps them equally much.

\paras{Do humans and LLMs find the same items difficult?}

To answer this, Table~\ref{tab:human-difficulty-main} groups items by how many of the three human judgments were correct. Model accuracy generally rises with human agreement, so the two notions of difficulty are related. The pattern is not perfectly clear for smaller models, however. Interestingly, GPT-5.6 Sol (max) manages to solve 72.5\% of the 43 items missed by all three human judgments. We did manually inspect these instances, but could not pinpoint a single cause. The model simply seems better at generating, selecting and evaluating possible answer candidates.

\section{Discussion}

\begin{table}[t]
\centering
\footnotesize
\begin{tabularx}{\columnwidth}{@{}>{\raggedright\arraybackslash}X rrrr@{}}
\toprule
\textbf{Category} & \textbf{Fietje} & \textbf{Luna} & \textbf{Sol} & \textbf{Human} \\
\midrule
Empty & 1 & 17 & 7 & 47 \\
Completely incorrect & 21 & 0 & 0 & 0 \\
Repeats (part of) a clue & 41 & 0 & 0 & 0 \\
Fits only one clue & 90 & 7 & 3 & 15 \\
Fits only two clues & 35 & 16 & 13 & 16 \\
Inflection variant & 0 & 4 & 2 & 0 \\
Correct alternative & 0 & 3 & 3 & 1 \\

\midrule
Total & 188 & 47 & 28 & 79 \\
\bottomrule
\end{tabularx}
\caption{Manual categories for answers initially marked incorrect in zero-shot canonical-order Fietje~2 Instruct, GPT-5.6 runs with max reasoning, and one full-test human annotator.}
\label{tab:error-annotation-categories-main}
\end{table}

\paragraph{What types of mistakes are made?}
We manually categorize the answers initially marked incorrect from one low-performing model (Fietje~2 Instruct), two high-performing models (GPT-5.6 Luna and Sol), and the best full-test human annotator. Usually, when the human annotator or good models make a mistake, their answers still fit one or two clues but fail on the rest (Table~\ref{tab:error-annotation-categories-main}). Luna and Sol even each produced three correct alternatives, and the human annotator did this once.\footnote{As noted before, these responses are counted as correct in the final benchmark. Appendix Table~\ref{tab:plausible-alternatives} lists all accepted alternatives and explains why they are acceptable.}
\textit{Empty} is also an interesting category. For the human annotator, this means that they could not come up with an answer they considered good enough and left the line empty. For a model, it means that they did not come up with an answer before running out of reasoning or generation tokens. On inspection, we saw that the models mostly looped through the same answers, and were no longer making meaningful progress when they ran out of tokens.
Fietje is clearly not up to the task: it often repeats all or part of a clue instead of proposing a new answer, or comes up with an answer that only fits a single clue.

\begin{table}[t]
\centering
\small
\setlength{\tabcolsep}{5pt}
\begin{tabular}{lrr}
\toprule
\textbf{Feature} & \textbf{Models} & \textbf{Human} \\
\midrule
Answer frequency & $\mathbf{.45}$ & $\mathbf{.34}$ \\
Average clue frequency & $.10$ & $\mathbf{.19}$ \\
Minimum clue frequency & $-.07$ & $.01$ \\
Average clue length & $.11$ & $.00$ \\
Answer length & $-.13$ & $-.14$ \\
Named-entity clues & $-.03$ & $-.01$ \\
Original disambiguation entries & $\mathbf{.38}$ & $\mathbf{.25}$ \\
Wikipedia page views & $\mathbf{.25}$ & $\mathbf{.28}$ \\
\bottomrule
\end{tabular}
\caption{Item-level Spearman correlations with mean model accuracy and human success. Bold values are significant at $p<.05$.}
\label{tab:benchmark-characterization-main}
\end{table}

\begin{figure}[t]
\centering
\includegraphics[width=\columnwidth]{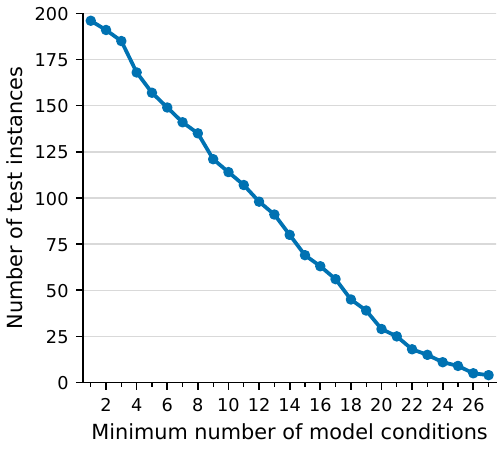}
\caption{Number of test instances solved by at least the number of model conditions on the horizontal axis. A model condition solves an instance if it produces an accepted answer in at least one of the six clue orders. 
Counts are cumulative, so an instance solved by, for example, ten models also contributes to every point from one to ten.}
\label{fig:model-solve-coverage}
\end{figure}

\paragraph{Are the instances discriminative?}
Figure~\ref{fig:model-solve-coverage} shows how many instances are solved by at least a given number of model conditions, counting an instance as solved when a condition succeeds in any clue order. Only four instances are never solved, while also only four are solved in all 27 conditions. We consider this scarcity at both extremes a desirable property for a benchmark: even though the benchmark is relatively small, many of its instances play a role in distinguishing one model from another.

\paragraph{What makes an item difficult?}
Table~\ref{tab:benchmark-characterization-main} relates model and human accuracy to simple item characteristics.\footnote{Appendix~\ref{app:characterization-details} describes how each feature is computed.} Frequencies are estimated from a large Dutch news corpus \citep{goldhahn-etal-2012-building}.
As perhaps expected, familiar answers (checked by answer frequency and page views) and disambiguation pages with more entries are easier for both models and humans, while most other characteristics have little association with accuracy. The full breakdown is shown in Appendix Table~\ref{tab:benchmark-characterization}.

\vspace{-0.1cm}
\section{Conclusion}
\vspace{-0.1cm}

In this paper, we introduced the Dutch LLM benchmark \emph{The Shared Word} (\benchmark{}), which combines four properties that rarely occur together in language-model evaluation: it is language-specific, open-vocabulary, deterministically scored, and fully curated by humans. The task is simple to describe but not easy to solve. A model receives three clues for different senses or uses of one Dutch word and has to generate that word. 

The results show both headroom and clear progress. Most models in our selection struggle, and the strongest open-weight checkpoint reaches only 39.3\%. GPT-4o and GPT-4.1 score within the range observed for our human annotators, while GPT-5.6 Luna and Sol score substantially higher. The ablations also confirm the intended structure of the task: one clue is rarely enough, two help considerably, and all three give the strongest result.

The benchmark itself is deliberately Dutch, since its lexical relations do not survive translation. The construction process is more general, though. Other languages with similar resources could construct a similar benchmark by following our recipe, though it would require careful manual curation. We hope this work encourages more benchmarks that test language-specific knowledge directly without giving up open generation or clear evaluation.

\section*{Limitations}

As indicated in the method, our benchmark is relatively small. We chose 200 carefully curated test items over a larger set of weaker puzzles. That choice limits statistical power and perhaps the ability to distinguish between models. We do believe this trade-off between quality and size is worth it in this case.

Our benchmark is also just for a single language. This is entirely on purpose: we believe we need more carefully curated, language-specific datasets instead of large, automatic datasets across multiple languages that are not manually verified. The same construction process could be applied to other languages with sufficient disambiguation resources. We sincerely hope future work will look into this.

Dataset construction required extensive manual curation. It involved judgments about familiarity, difficulty, and whether an answer would be unique. Only two curators were involved in these decisions. This introduces some subjectivity into item selection, as is unavoidable when curating this kind of puzzle benchmark. A different team might therefore select a somewhat different final set.

Since the task is derived from Wikipedia, pretrained models may benefit from associations acquired from that source, but the low performance of most evaluated models shows that source exposure alone is not enough. Because the task is open-vocabulary, valid alternatives cannot be guaranteed to be exhaustively known in advance. Therefore, we tried our best to identify such alternatives during error analysis, and updated all scores in the paper taking them into account.

Finally, TSW-NL is mainly a test of Dutch lexical knowledge, candidate generation, and evidence integration. In particular, while it tests open-vocabulary generation, it does not test generation capabilities beyond just words. We see it as a nice addition to the current evaluation landscape, but not, of course, as a superior benchmark in testing general Dutch capabilities.

\section{Acknowledgments}

We thank the Center for Information Technology of the University of Groningen for their support and for providing access to the Hábrók high performance computing cluster. We also thank all our colleagues of the GroNLP group for the discussions and useful feedback. Finally, we thank all the annotators that took part in this study.

\bibliography{custom}


\appendix

\section{Related Language-Specific Benchmarks}

Tables~\ref{tab:related-benchmarks}, \ref{tab:related-benchmark-suites}, and~\ref{tab:dutch-benchmark-tasks} apply the same four criteria to the benchmarks included in our literature review. For benchmarks containing multiple tasks, \mixed{} means that only some tasks have the property. Our definition of single-language is stricter than simply evaluating text in the target language: translations of another benchmark do not pass SL.

\section{Human Evaluation Protocol}
\label{app:human}

\para{Participants and research ethics}

Participants were recruited through our personal network and received no compensation. They were all native speakers of Dutch. Before participating, they were informed that their anonymized answers would be used in research and gave consent. The task involved solving ordinary lexical puzzles and did not collect personal or sensitive information. No ethical clearance from our institution was therefore deemed necessary. 

\para{Participant instructions}

Participants received the same instructions as the LLMs, except for the instructions about JSON output. They were not shown example puzzles. There was no time limit. They were not allowed to use browsing or any other external resource.

\section{Licenses}
\label{app:licenses}

The title filter uses an archived lemma list derived from Open Dutch WordNet \citep{postma-etal-2016-open}, distributed under CC BY-SA 4.0. The candidate clues were derived from Dutch Wikipedia text. At the time of collection, this material was also available under CC BY-SA 4.0. We will release our benchmark under a Apache 2.0 license.

\section{Example Test Items}

Table~\ref{tab:test-examples} gives ten test items, with English translations for readability.

\section{Correct Alternative Answers}

Table~\ref{tab:plausible-alternatives} lists the seven items for which error annotation identified a second correct answer. Both answers are accepted by the final benchmark scorer. We added an explanation for why the alternative answer should be accepted. The final three answers can hardly be considered \emph{different} answers: they are plural or verbal versions of the original answer.

\begin{table*}[t]
\centering
\small
\setlength{\tabcolsep}{2.4pt}
\renewcommand{\arraystretch}{1.02}
\begin{tabularx}{\textwidth}{@{}>{\raggedright\arraybackslash}p{0.18\textwidth} >{\raggedright\arraybackslash}p{0.085\textwidth} >{\raggedright\arraybackslash}X cccc@{}}
\toprule
\textbf{Benchmark} & \textbf{Language} & \textbf{Brief description} & \textbf{SL} & \textbf{OV} & \textbf{HV} & \textbf{DE} \\
\midrule
\benchmark{} (ours) & Dutch & Three clues identify one ambiguous Dutch word & \pass & \pass & \pass & \pass \\
\midrule
JAWAHER \citep{magdy-etal-2025-jawaher} & Arabic/\newline English & Multidialectal proverb translation and explanation (0/3 tasks pass) & \mixed & \pass & \pass & \fail \\
Nahw \citep{mubarak-etal-2026-nahw} & Arabic & Grammar analysis, correction, and explanation (2/4 tasks pass) & \pass & \mixed & \pass & \mixed \\
ALARB \citep{shairah-etal-2025-alarb} & Arabic & Legal verdict prediction and statute identification (0/6 tasks pass) & \pass & \mixed & \fail & \mixed \\
SAHM \citep{elbadry-etal-2026-sahm} & Arabic & Financial and Shari'ah-compliant reasoning (0/7 tasks pass) & \pass & \mixed & \pass & \mixed \\
BALSAM \citep{almatham-etal-2025-balsam} & Arabic & Arabic understanding and generation platform (0/78 tasks pass) & \mixed & \mixed & \fail & \mixed \\
BanglaSocialBench \citep{sijan-etal-2026-banglasocialbench} & Bangla & Address terms, kinship, and social customs & \pass & \fail & \pass & \pass \\
CDQA \citep{xu-etal-2025-llms} & Chinese & Time-sensitive, human-verified short-answer questions collected from the Chinese web & \pass & \pass & \pass & \fail \\
MEDFACT Response Checking \citep{chen-etal-2025-medfact} & Chinese & Evidence-based fact checking of LLM-generated Chinese medical responses & \pass & \fail & \pass & \pass \\
MedFact Veracity and Localization \citep{he-etal-2026-medfact} & Chinese & Chinese medical veracity classification and error localization (0/2 tasks pass) & \pass & \mixed & \pass & \mixed \\
CHENGYU-BENCH \citep{fu-etal-2025-chengyu} & Chinese & Chinese idiom connotation, contextual appropriateness, and open-cloze use (1/3 tasks pass) & \pass & \mixed & \pass & \pass \\
CTM \citep{wang-etal-2026-benchmarking} & Chinese & Chinese dynastic chronology QA and timeline-order reasoning (0/9 tasks pass) & \pass & \mixed & \pass & \pass \\
LexGenius \citep{liu-etal-2026-lexgenius} & Chinese & Expert-level Chinese legal knowledge and reasoning across 11 MCQ tasks (0/11 tasks pass) & \pass & \fail & \pass & \pass \\
LAiW \citep{dai-etal-2025-laiw} & Chinese & Chinese legal syllogism tasks from retrieval to legal application (0/14 tasks pass) & \pass & \mixed & \mixed & \mixed \\
Danish Semantic Reasoning \citep{pedersen-etal-2024-towards} & Danish & Curated Danish lexical-semantic inference, synonymy, relatedness, and word-sense tasks (0/25 tasks pass) & \pass & \fail & \pass & \pass \\
CrowS-Pairs-NL \citep{van-der-weide-etal-2026-crows} & Dutch & Dutch stereotype-bias preference pairs adapted from English and crowdsourced locally & \fail & \fail & \pass & \fail \\
LCHAIM \citep{malul-etal-2025-lchaim} & Hebrew & Long-context Hebrew natural-language inference translated from ConTRoL & \fail & \fail & \pass & \pass \\
Icelandic Linguistic Benchmark \citep{armannsson-etal-2025-icelandic} & Icelandic & Icelandic syntax, morphology, anaphora, and translation diagnostics (5/21 tasks pass) & \mixed & \mixed & \pass & \mixed \\
\bottomrule
\end{tabularx}
\caption{Recent LLM benchmarks developed for non-English languages, part 1. \textbf{SL}: created for one target language rather than translated from another benchmark or evaluated in parallel target-language and English versions. \textbf{OV}: the task requires generating an answer rather than selecting from fixed candidates. \textbf{HV}: construction includes substantive human checking, curation and validation. \textbf{DE}: answers are deterministically evaluated using a fixed answer or rule without needing an LLM judge, human rating, or fuzzy metric. For multi-task benchmarks, \pass{}, \mixed{}, and \fail{} mean that all, some, or none of the constituent tasks pass, respectively.}
\label{tab:related-benchmarks}
\end{table*}

\begin{table*}[t]
\centering
\small
\setlength{\tabcolsep}{2.4pt}
\renewcommand{\arraystretch}{1.02}
\begin{tabularx}{\textwidth}{@{}>{\raggedright\arraybackslash}p{0.18\textwidth} >{\raggedright\arraybackslash}p{0.085\textwidth} >{\raggedright\arraybackslash}X cccc@{}}
\toprule
\textbf{Benchmark} & \textbf{Language} & \textbf{Brief description} & \textbf{SL} & \textbf{OV} & \textbf{HV} & \textbf{DE} \\
\midrule
ITALIC \citep{seveso-etal-2025-italic} & Italian & Italian culture, history, civic knowledge, and language questions from public examinations (0/12 tasks pass) & \pass & \fail & \pass & \pass \\
SOBACO \citep{yamamoto-etal-2025-bias} & Japanese & Paired questions testing Japanese age, gender, hierarchy, and workplace norms & \pass & \fail & \pass & \pass \\
JPharmaBench \citep{ono-etal-2025-japanese} & Japanese & Japanese pharmaceutical licensing, terminology normalization, and consistency checking (0/3 tasks pass) & \mixed & \fail & \pass & \pass \\
Japanese Financial Benchmark \citep{hirano-2024-construction} & Japanese & Japanese financial sentiment, knowledge, audit, and professional-exam tasks (0/5 tasks pass) & \pass & \fail & \fail & \pass \\
JMedBench \citep{jiang-etal-2025-jmedbench} & Japanese & Japanese biomedical MCQA, translation, NER, classification, and similarity tasks (0/5 tasks pass) & \mixed & \mixed & \mixed & \mixed \\
KBL \citep{kimyeeun-etal-2024-developing} & Korean & Korean legal knowledge and reasoning & \pass & \fail & \pass & \pass \\
KorNAT \citep{lee-etal-2024-kornat} & Korean & Social values and common knowledge & \pass & \fail & \pass & \pass \\
Nunchi-Bench \citep{kim-lee-2025-nunchi} & Korean/\newline English & Korean superstition knowledge, advice, and interpretation (0/3 tasks pass) & \fail & \mixed & \pass & \mixed \\
CLIcK \citep{kim-etal-2024-click} & Korean & Korean cultural and linguistic intelligence across 11 exam- and textbook-based categories (0/11 tasks pass) & \pass & \fail & \pass & \pass \\
PRONE \citep{thapa-etal-2025-probing} & Nepali & Nepali proverb meaning selection and thematic classification (0/2 tasks pass) & \pass & \fail & \pass & \pass \\
NorEval \citep{mikhailov-etal-2025-noreval} & Norwegian & Bokm\r{a}l and Nynorsk understanding, generation, idiom, and knowledge tasks (3/24 tasks pass) & \mixed & \mixed & \pass & \mixed \\
TAAROFBENCH \citep{sadr-etal-2025-politely} & Persian/\newline English & Open-ended role play about Persian taarof & \fail & \pass & \pass & \fail \\
DIVANBENCH \citep{sakhaeirad-etal-2026-unmasking} & Persian & Superstitions, customs, and social etiquette (0/3 tasks pass) & \pass & \fail & \fail & \fail \\
FFE-Hallu \citep{hosseini-etal-2026-ffe} & Persian/\newline English & Idiom/proverb generation, detection, and translation (0/3 tasks pass) & \mixed & \mixed & \pass & \mixed \\
Persian LLM Evaluation \citep{hosseinbeigi-etal-2025-advancing} & Persian & Persian language understanding, academic knowledge, and culture tasks with MCQ and open-ended formats (1/9 tasks pass) & \mixed & \mixed & \pass & \mixed \\
SweSAT-1.0 \citep{kurfali-etal-2025-swesat} & Swedish & Swedish university entrance-exam verbal and quantitative reasoning (0/6 tasks pass) & \pass & \fail & \pass & \pass \\
TimeRes \citep{demir-etal-2026-timeres} & Turkish & Temporal interpretation and date calculation (0/4 tasks pass) & \pass & \pass & \fail & \pass \\
TurkBench \citep{toraman-etal-2026-turkbench} & Turkish & Turkish knowledge, reasoning, language, safety, and cultural evaluation tasks (0/21 tasks pass) & \mixed & \mixed & \pass & \mixed \\
CETVEL \citep{er-etal-2026-cetvel} & Turkish & Native, translated, and multilingual Turkish understanding, generation, and culture tasks (0/23 tasks pass) & \mixed & \mixed & \mixed & \mixed \\
\bottomrule
\end{tabularx}
\caption{Continuation of Table~\ref{tab:related-benchmarks}.}
\label{tab:related-benchmark-suites}
\end{table*}

\begin{table*}[t]
\centering
\small
\setlength{\tabcolsep}{2.4pt}
\renewcommand{\arraystretch}{1.02}
\begin{tabularx}{\textwidth}{@{}>{\raggedright\arraybackslash}p{0.18\textwidth} >{\raggedright\arraybackslash}p{0.085\textwidth} >{\raggedright\arraybackslash}X cccc@{}}
\toprule
\textbf{Benchmark} & \textbf{Language} & \textbf{Brief description} & \textbf{SL} & \textbf{OV} & \textbf{HV} & \textbf{DE} \\
\midrule
\benchmark{} (ours) & Dutch & Three clues identify one ambiguous Dutch word & \pass & \pass & \pass & \pass \\
\midrule
\multicolumn{7}{@{}l}{\textit{EuroEval Dutch main-score tasks \citep{saattrup-nielsen-etal-2025-encoder}}} \\
\addlinespace[1pt]
CoNLL-NL & Dutch & Named-entity recognition & \pass & \fail & \pass & \pass \\
DBRD & Dutch & Sentiment classification & \pass & \fail & \pass & \pass \\
Duidelijke Taal & Dutch & Text simplification & \pass & \pass & \pass & \fail \\
HellaSwag-NL & Dutch & Multiple-choice common-sense reasoning & \fail & \fail & \fail & \pass \\
MMLU-NL & Dutch & Multiple-choice knowledge & \fail & \fail & \fail & \pass \\
ScaLA-NL & Dutch & Linguistic acceptability & \pass & \fail & \fail & \pass \\
SQuAD-NL & Dutch & Extractive question answering & \fail & \pass & \pass & \pass \\
WikiLingua-NL & Dutch & Abstractive summarization & \fail & \pass & \pass & \fail \\
\midrule
\multicolumn{7}{@{}l}{\textit{BEIR-NL \citep{lotfi-etal-2025-beir}}} \\
\addlinespace[1pt]
BEIR-NL & Dutch & Automatically translated information retrieval & \fail & \fail & \fail & \pass \\
\midrule
\multicolumn{7}{@{}l}{\textit{DUMB tasks \citep{de-vries-etal-2023-dumb}}} \\
\addlinespace[1pt]
COPA-NL & Dutch & Causal reasoning & \fail & \fail & \pass & \pass \\
DALC & Dutch & Abusive-language classification & \pass & \fail & \pass & \pass \\
DBRD & Dutch & Sentiment classification & \pass & \fail & \pass & \pass \\
DPR & Dutch & Pronoun resolution & \pass & \fail & \pass & \pass \\
Lassy POS & Dutch & Part-of-speech tagging & \pass & \fail & \pass & \pass \\
SICK-NL & Dutch & Natural-language inference & \fail & \fail & \pass & \pass \\
SoNaR-1 NER & Dutch & Named-entity recognition & \pass & \fail & \pass & \pass \\
SQuAD-NL & Dutch & Extractive question answering & \fail & \pass & \pass & \pass \\
WiC-NL & Dutch & Word-sense discrimination & \pass & \fail & \pass & \pass \\
\bottomrule
\end{tabularx}
\caption{Dutch tasks in EuroEval, DUMB, and BEIR-NL, compared with \benchmark{}. BEIR-NL is translated automatically from the English BEIR datasets, ranks documents from a fixed corpus, reports deterministic retrieval metrics, and does not include substantive human verification of its Dutch construction. Column definitions and symbols follow Table~\ref{tab:related-benchmarks}.}
\label{tab:dutch-benchmark-tasks}
\end{table*}


\begin{table*}[t]
\centering
\small
\resizebox{\textwidth}{!}{%
\begin{tabular}{@{}llll@{}}
\toprule
\textbf{Clue 1} & \textbf{Clue 2} & \textbf{Clue 3} & \textbf{Answer} \\
\midrule
Een plant & Een gemeenschap van mensen & Een term uit taalkunde & Stam \\
\textit{A plant} & \textit{A community of people} & \textit{A linguistic term} & \\
\addlinespace[5pt]
Een Nederlandse televisiepresentator & Vuur & Een Nederlandse brouwerij & Brand \\
\textit{A Dutch television presenter} & \textit{Fire} & \textit{A Dutch brewery} & \\
\addlinespace[5pt]
Een kleur & Een strafkaart & Een politieke richting & Rood \\
\textit{A colour} & \textit{A penalty card} & \textit{A political orientation} & \\
\addlinespace[5pt]
Een reeks tekens in een postadres & Een versleuteling van gegevens & Een computertaal & Code \\
\textit{A string of characters in a postal address} & \textit{An encryption of data} & \textit{A programming language} & \\
\addlinespace[5pt]
Een fossiele hars & Een kleur & Een vrouwelijke voornaam & Amber \\
\textit{A fossil resin} & \textit{A colour} & \textit{A female given name} & \\
\addlinespace[5pt]
Een houtsoort & Een onderdeel van een kerkgebouw & Een Nederlandse televisiekok & Beuk \\
\textit{A type of wood} & \textit{Part of a church building} & \textit{A Dutch television chef} & \\
\addlinespace[5pt]
Een onderdeel van een schip & Een onderdeel van een molen & Een type vloerbedekking & Zeil \\
\textit{Part of a ship} & \textit{Part of a windmill} & \textit{A type of floor covering} & \\
\addlinespace[5pt]
Een muziekinstrument & Een rookgerei & Een waterafvoer & Pijp \\
\textit{A musical instrument} & \textit{A smoking implement} & \textit{A drain} & \\
\addlinespace[5pt]
Een Nederlandse volksvertegenwoordiging & Een onderdeel van een woning & Onderdeel van het hart en vaatstelsel & Kamer \\
\textit{A Dutch representative body} & \textit{Part of a home} & \textit{Part of the cardiovascular system} & \\
\addlinespace[5pt]
Een plant & Een datastructuur & Een staaf horizontaal over een weg & Boom \\
\textit{A plant} & \textit{A data structure} & \textit{A horizontal barrier across a road} & \\
\bottomrule
\end{tabular}%
}
\caption{Ten example test items. English translations are shown in italics and are not part of the benchmark input.}
\label{tab:test-examples}
\end{table*}


\begin{table*}[t]
\centering
\footnotesize
\setlength{\tabcolsep}{4pt}
\begin{tabularx}{\textwidth}{@{}
    >{\raggedright\arraybackslash}p{4.4cm}
    >{\raggedright\arraybackslash}X
    >{\raggedright\arraybackslash}X
    >{\raggedright\arraybackslash}p{1.4cm}
    >{\raggedright\arraybackslash}p{1.6cm}
    >{\raggedright\arraybackslash}p{1.65cm}@{}}
\toprule
\textbf{Clue 1} & \textbf{Clue 2} & \textbf{Clue 3} & \textbf{Gold} & \textbf{Alternative} & \textbf{Who} \\
\midrule
Het terrein van beroepsmatig toegepaste kennis & Een term in sport & Het geheel van arbeidstaken & Discipline & Vak & Luna, Sol, human \\
\textit{A field of professionally applied knowledge} & \textit{A term in sport} & \textit{The set of work duties} & & & \\
\addlinespace[4pt]
\multicolumn{6}{@{}l@{}}{\textbf{Why accept?} \emph{Vak} perfectly fits clues 1 and 3. The second one is too vague to really rule it out: \emph{vak} is also a term in sport.} \\
\midrule
Een automodel & Een type vliegtuig & Een wild dier & Eend & Jaguar & Luna \\
\textit{A car model} & \textit{A type of aircraft} & \textit{A wild animal} & & & \\
\addlinespace[4pt]
\multicolumn{6}{@{}l@{}}{\textbf{Why accept?} \emph{Jaguar} even fits the first and third clues better than \emph{Eend} and both are relatively obscure aircraft.} \\
\midrule
Een eenheid & Een mannelijke voornaam & Een oude computer & Pascal & Mark & Sol \\
\textit{A unit} & \textit{A male given name} & \textit{An old computer} & & & \\
\addlinespace[4pt]
\multicolumn{6}{@{}l@{}}{\textbf{Why accept?} Both general clues fit. \emph{Mark} is a very obscure old computer, but Sol is still technically correct.} \\
\midrule
Een verzorger & Een lid van een kloosterorde & Een familielid & Zuster & Broeder & Luna, Sol \\
\textit{A caregiver} & \textit{A member of a religious order} & \textit{A family member} & & & \\
\addlinespace[4pt]
\multicolumn{6}{@{}l@{}}{\textbf{Why accept?} \emph{Broeder} and \emph{Zuster} are old-fashioned words for brother and sister, and both fit the descriptions.} \\
\multicolumn{6}{@{}l@{}}{\hspace{1.75cm} This was an oversight during data curation.} \\
\midrule
Een temperatuurschaal & Een academisch concept & Een wiskundig concept & Graad & Graden & GPT-4o, Ministral \\
\textit{A temperature scale} & \textit{An academic concept} & \textit{A mathematical concept} & & & \\
\addlinespace[4pt]
\multicolumn{6}{@{}l@{}}{\textbf{Why accept?} \emph{Graden} is the regular plural of \emph{graad} and fits the same three clues.} \\
\midrule
Een term uit muziektheorie & Een opmerking uit een tekst & Een vrucht & Noot & Noten & Aya, EuroLLM, Qwen3, Salamandra \\
\textit{A term from music theory} & \textit{A note in a text} & \textit{A fruit} & & & \\
\addlinespace[4pt]
\multicolumn{6}{@{}l@{}}{\textbf{Why accept?} \emph{Noten} is the regular plural of \emph{noot} and fits the same three clues.} \\
\midrule
Het zoeken van kandidaten & Een onderdeel van evolutie & Een grafische gebruikersomgeving handeling & Selectie & Selecteren & Aya, ChocoLlama, Llama 3, Llama 3.1, Qwen3, Salamandra \\
\textit{The search for candidates} & \textit{A component of evolution} & \textit{A graphical user interface action} & & & \\
\addlinespace[4pt]
\multicolumn{6}{@{}l@{}}{\textbf{Why accept?} \emph{Selecteren} is the corresponding verb and fits the three described processes.} \\
\bottomrule
\end{tabularx}
\caption{Items for which a model or human annotator produced a correct alternative to the gold answer. These alternatives are included in the accepted-answer sets used for all reported scores.}
\label{tab:plausible-alternatives}
\end{table*}
\FloatBarrier

\begin{figure*}[p]
\centering
\includegraphics[width=0.92\textwidth]{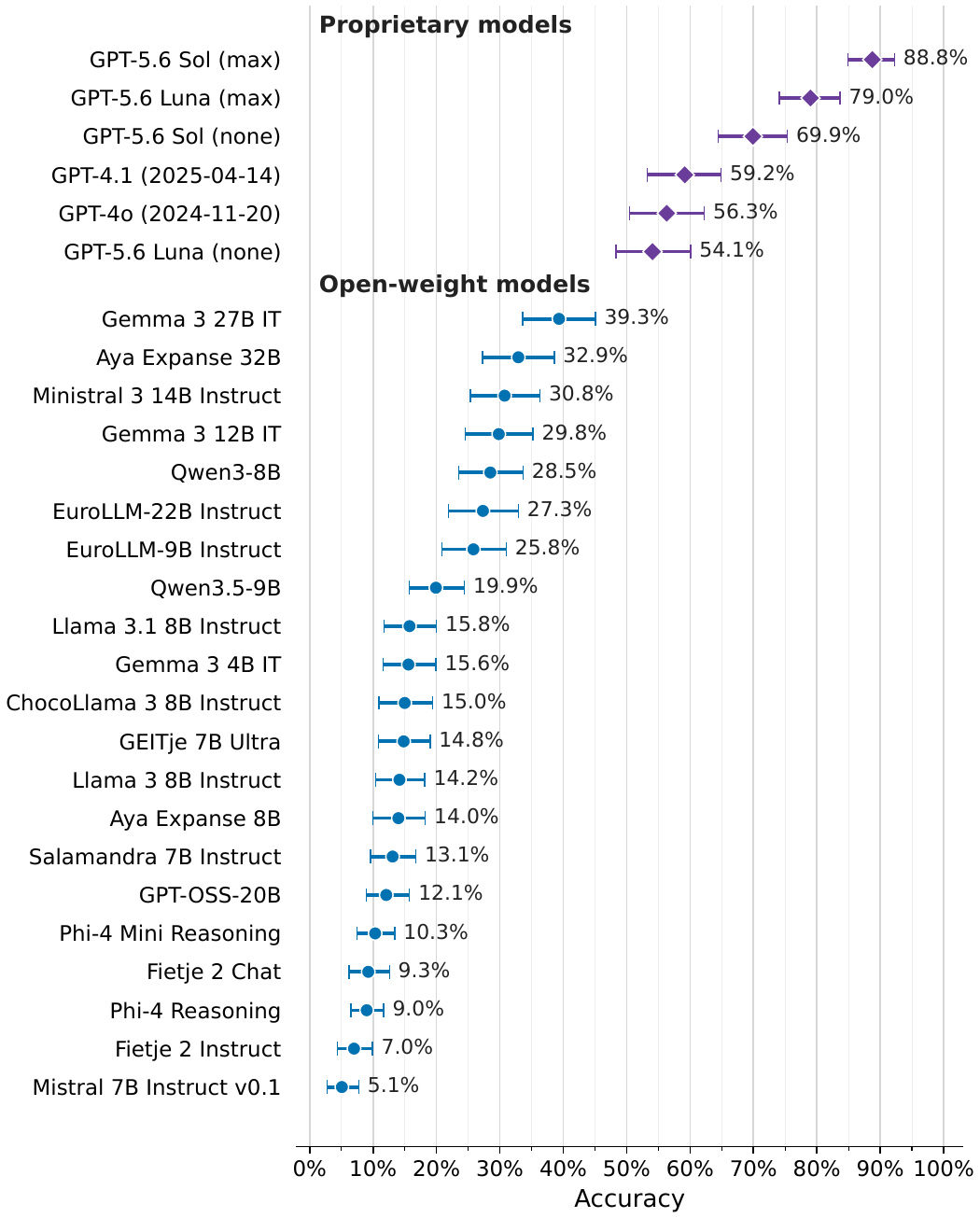}
\caption{Test accuracy for all models included in the evaluation, averaged over the six permutations of the three clues. Dots show the mean accuracy and lines show 95\% confidence intervals obtained by bootstrapping the 200 test items while keeping the six clue orders for each item together.}
\label{fig:main-results-forest}
\end{figure*}

\FloatBarrier

\section{Prompt Templates}
\label{app:prompts}

\noindent\textbf{System message}\par\smallskip
{\nolinenumbers
\noindent\fbox{\begin{minipage}{0.94\columnwidth}
\small\ttfamily
Je bent een taalexpert die Nederlandse taalkundige raadsels oplost. \\

Gebruik geen browser, webzoekopdracht, externe bron of tool. \\

Los het raadsel alleen op met je interne taalkennis.\\

Geef het antwoord altijd in het Nederlands, niet in het Engels.
\end{minipage}}
\par}

\medskip\noindent\textbf{User message}\par\smallskip
{\nolinenumbers
\noindent\fbox{\begin{minipage}{0.94\columnwidth}
\small\ttfamily
Los het volgende taalkundige raadsel op. Het bevat drie aanwijzingen, en het antwoord is één woord dat daarop van toepassing is.

\medskip
Geef alleen het antwoord zelf in JSON. Geen uitleg, geen markdown, geen extra sleutels.

\medskip
Het antwoord moet Nederlands zijn, niet Engels.

\medskip
Het JSON-formaat moet exact zijn: \{"answer": "<antwoord>"\}.

\medskip
Als er meerdere geldige opties zijn, kies dan je beste gok.

\medskip
Vraag: \{puzzel\}. Wat is het?\\
Antwoord:
\end{minipage}}
\par}

\medskip

\section{Experimental Details}
\label{app:reproducibility}

\begin{table}[t]
\centering
\small
\begin{tabularx}{\columnwidth}{>{\raggedright\arraybackslash}X rrr}
\toprule
\textbf{Model} & \textbf{1} & \textbf{2} & \textbf{3} \\
\midrule
Fietje 2 Instruct & 2.0 & 5.3 & 6.0 \\
Fietje 2 Chat & 2.7 & 7.0 & 9.5 \\
GEITje 7B Ultra & 3.7 & 9.5 & 15.0 \\
ChocoLlama 3 8B Instruct & 4.3 & 9.2 & 15.5 \\
\midrule
Mistral 7B Instruct v0.1 & 2.2 & 3.2 & 3.5 \\
Llama 3 8B Instruct & 5.5 & 11.7 & 19.5 \\
Salamandra 7B Instruct & 4.8 & 9.5 & 14.0 \\
Aya Expanse 8B & 4.2 & 11.0 & 13.5 \\
Aya Expanse 32B & 8.2 & 23.0 & 33.5 \\
Llama 3.1 8B Instruct & 5.2 & 10.3 & 14.0 \\
Gemma 3 4B IT & 3.7 & 10.7 & 16.0 \\
Gemma 3 12B IT & 8.2 & 19.0 & 31.0 \\
Gemma 3 27B IT & 9.2 & 26.3 & 39.0 \\
EuroLLM-9B Instruct & 6.2 & 16.7 & 27.5 \\
EuroLLM-22B Instruct & 5.5 & 17.7 & 26.5 \\
Ministral 3 14B Instruct & 6.7 & 20.0 & 31.5 \\
\midrule
Qwen3-8B & 6.8 & 21.2 & 26.0 \\
Qwen3.5-9B & 5.4 & 15.5 & 22.5 \\
Phi-4 Mini Reasoning & 1.7 & 4.3 & 11.5 \\
Phi-4 Reasoning & 0.0 & 3.2 & 9.5 \\
GPT-OSS-20B & 0.3 & 9.0 & 12.0 \\
\midrule
GPT-4o (2024-11-20) & 11.0 & 35.5 & 57.5 \\
GPT-4.1 (2025-04-14) & 11.3 & 39.8 & 60.0 \\
GPT-5.6 Luna (none) & 9.0 & 34.2 & 54.5 \\
GPT-5.6 Luna (max) & 9.7 & 57.7 & 78.0 \\
GPT-5.6 Sol (none) & 9.7 & 43.0 & 68.0 \\
GPT-5.6 Sol (max) & 10.3 & 63.2 & 87.5 \\
\bottomrule
\end{tabularx}
\caption{Accuracy (\%) by number of clues. Column 1 is the average of the single-clue conditions 1, 2, and 3. Column 2 averages the two-clue conditions 1+2, 1+3, and 2+3. Finally, column 3 uses all three clues in canonical order.}
\label{tab:ablation-results}
\end{table}

\begin{table}[!t]
  \centering
  \small
  \setlength{\tabcolsep}{2pt}
  \resizebox{\columnwidth}{!}{%
  \begin{tabular}{lrrrrrr}
  \toprule
  \textbf{Model} & \textbf{0} & \textbf{1} & \textbf{3} & \textbf{5} & \textbf{10} & \textbf{22} \\
  \midrule
  Fietje 2 Instruct & 6.0 & \textbf{7.0} & \textbf{7.0} & 6.0 & 6.5 & 3.5 \\
  Fietje 2 Chat & \textbf{9.5} & 7.5 & 7.0 & 6.5 & 6.0 & 8.5 \\
  GEITje 7B Ultra & 15.0 & 11.0 & 16.5 & 15.0 & \textbf{17.5} & 15.5 \\
  ChocoLlama 3 8B Instruct & 15.5 & 15.0 & 17.0 & \textbf{17.5} & 16.5 & 16.5 \\
  \midrule
  Mistral 7B Instruct v0.1 & 3.5 & 10.5 & 9.5 & 10.0 & \textbf{11.0} & \textbf{11.0} \\
  Llama 3 8B Instruct & 19.5 & 16.5 & 18.5 & 19.0 & \textbf{20.5} & 19.0 \\
  Salamandra 7B Instruct & 14.0 & 14.5 & \textbf{15.5} & 13.5 & 13.5 & 10.5 \\
  Aya Expanse 8B & 13.5 & \textbf{19.5} & 19.0 & 16.5 & 17.5 & 17.0 \\
  Aya Expanse 32B & 33.5 & 35.5 & \textbf{36.5} & 35.0 & 34.0 & 33.0 \\
  Llama 3.1 8B Instruct & 14.0 & 17.5 & 20.5 & 21.0 & 20.0 & \textbf{23.5} \\
  Gemma 3 4B IT & 16.0 & 18.0 & 17.0 & \textbf{22.0} & 21.0 & 18.0 \\
  Gemma 3 12B IT & 31.0 & 31.0 & 32.5 & 34.0 & 33.0 & \textbf{35.0} \\
  Gemma 3 27B IT & 39.0 & 41.0 & 41.0 & 40.5 & 39.5 & \textbf{44.5} \\
  EuroLLM-9B Instruct & \textbf{27.5} & 18.0 & 19.0 & 17.5 & 18.0 & 20.5 \\
  EuroLLM-22B Instruct & \textbf{26.5} & 24.5 & 21.5 & 20.5 & 22.0 & 24.5 \\
  Ministral 3 14B Instruct & 31.5 & 33.0 & 31.5 & 33.0 & \textbf{35.5} & 34.5 \\
  \midrule
  Qwen3-8B & 26.0 & 24.0 & 22.0 & 23.5 & 23.5 & \textbf{27.0} \\
  Qwen3.5-9B & \textbf{22.5} & 19.5 & 13.5 & 15.5 & 14.5 & 14.0 \\
  Phi-4 Mini Reasoning & \textbf{11.5} & 7.5 & 8.5 & 9.0 & 8.5 & 9.0 \\
  Phi-4 Reasoning & \textbf{9.5} & 5.5 & 5.0 & 7.0 & 7.0 & 7.0 \\
  \bottomrule
  \end{tabular}%
  }
  \caption{Accuracy (\%) for a subset of models in one canonical-order run under each shot condition. Best result per model is shown in bold.}
  \label{tab:shot-selection}
  \end{table}

\begin{table*}[!t]
\centering
\small
\begin{tabular*}{\textwidth}{@{\extracolsep{\fill}}lcccc@{}}
\toprule
\textbf{Feature} & \textbf{All} & \textbf{Open} & \textbf{Prop.} & \textbf{Human} \\
\midrule
Answer frequency & $.45$ ($<.001$) & $.44$ ($<.001$) & $.42$ ($<.001$) & $.34$ ($<.001$) \\
Average clue frequency & $.10$ (.168) & $.05$ (.461) & $.17$ (.015) & $.19$ (.006) \\
Minimum clue frequency & $-.07$ (.329) & $-.10$ (.168) & $.05$ (.480) & $.01$ (.846) \\
Average clue length & $.11$ (.117) & $.12$ (.099) & $.10$ (.160) & $.00$ (.945) \\
Answer length & $-.13$ (.067) & $-.08$ (.283) & $-.17$ (.014) & $-.14$ (.052) \\
Named-entity clues & $-.03$ (.637) & $-.05$ (.449) & $.03$ (.707) & $-.01$ (.880) \\
Original disambiguation entries & $.38$ ($<.001$) & $.38$ ($<.001$) & $.31$ ($<.001$) & $.25$ ($<.001$) \\
Wikipedia page views & $.25$ ($<.001$) & $.25$ ($<.001$) & $.20$ (.005) & $.28$ ($<.001$) \\
\bottomrule
\end{tabular*}
\caption{Item-level Spearman correlations. Each cell shows $\rho$, with the two-sided $p$-value in parentheses. \textbf{All} denotes mean accuracy across the 27 model conditions, while \textbf{Open} and \textbf{Prop.} give the corresponding open-weight and proprietary means.}
\label{tab:benchmark-characterization}
\end{table*}

\para{Reproducibility} Each full clue-order evaluation contains 1,200 predictions: all six permutations for every test item. The development set was not used for optimizing performance, and it was only used for the shot items in Table~\ref{tab:shot-selection}. No settings or prompts were tuned on the test set. Local models were run with the same seed (42) and a maximum of 2,048 generated tokens on an NVIDIA A40 48GB or A100 80GB GPU. The settings not mentioned are left at their default values. 

We extract the requested JSON answer and use fixed fallbacks for a final-answer label or a plain first line when a model violates the output format. Our scoring strips surrounding whitespace, lowercases both strings, and otherwise requires exact equality with any accepted answer. Statistical resampling keeps the six orders of an item together. We use 10,000 item-bootstrap samples for accuracy and paired-difference intervals, 100,000 sign flips for paired model comparisons, and 10,000 within-item permutations for clue-order tests, all with the same seed as in the experiments. Holm correction is applied separately to the model-comparison and order-effect families. 

\para{OpenAI settings}
GPT models are queried through the OpenAI Chat Completions API. GPT-4o and GPT-4.1 are identified by dated snapshots (see the bottom of Table~\ref{tab:main-results}), but Sol and Luna did not have such a snapshot yet. We ran the experiments in July 2026. The experiments use either max or no reasoning effort and a maximum completion budget of 2,048 tokens, which includes hidden reasoning tokens.

\section{Benchmark Characterization}
\label{app:benchmark-characterization}
\label{app:characterization-details}

We test whether item difficulty is associated with simple lexical and source characteristics. We compute every feature once per test item and correlate it with item-level success using Spearman's $\rho$. For each model condition, correctness is averaged over the six clue orders. The model outcome is then the mean across all 27 conditions. Human success is the mean of the three human judgments for each item. Table~\ref{tab:benchmark-characterization} additionally separates the open-weight and proprietary model conditions. We process answers and clues with the Dutch spaCy model \texttt{nl\_core\_news\_sm}. Corpus frequencies are lemma counts from the one-million-sentence Dutch 2023 news corpus in the Leipzig Corpora Collection \citep{goldhahn-etal-2012-building}, normalized per million alphabetic tokens. The features are:

\begin{itemize}[itemsep=1pt,topsep=2pt]
    \item \textbf{Answer frequency}: the frequency of the answer lemma.
    \item \textbf{Average clue frequency}: the mean frequency of all lemmatized alphabetic non-stopwords across the three clues.
    \item \textbf{Minimum clue frequency}: the frequency of the least frequent clue in that set.
    \item \textbf{Average clue length}: the mean number of alphabetic tokens per clue, including stopwords.
    \item \textbf{Answer length}: the number of characters in the answer.
    \item \textbf{Named-entity clues}: the number of clues that contained at least one named entity.
    \item \textbf{Original disambiguation entries}: the number of linked entries on the answer's Dutch Wikipedia disambiguation page before filtering.
    \item \textbf{Wikipedia page views}: the mean stored 30-day page-view count of the three retained source entries.
\end{itemize}

Answer frequency has the strongest association with mean model accuracy ($\rho=.45$), followed by the number of entries on the original disambiguation page ($\rho=.38$) and Wikipedia page views ($\rho=.25$). The corresponding correlations are similar for open-weight and proprietary models. Minimum clue frequency, clue length, and the number of clues containing a named entity have little relation to mean model accuracy. These results suggest that familiar answers and richer source pages make an item easier, but that benchmark difficulty is not largely explained by the simple properties tested here.

\section{AI Assistance}

We used AI tools mainly for coding assistance, and a bit for proofreading. All AI-generated output was verified by the authors.

\end{document}